\documentclass[a4paper,fleqn]{cas-dc}

\usepackage[numbers]{natbib}
\usepackage{adjustbox}

\def\tsc#1{\csdef{#1}{\textsc{\lowercase{#1}}\xspace}}
\tsc{WGM}
\tsc{QE}
\tsc{EP}
\tsc{PMS}
\tsc{BEC}
\tsc{DE}

\let\printorcid\relax

\begin{document}
\let\WriteBookmarks\relax
\def\floatpagepagefraction{1}
\def\textpagefraction{.001}

\shorttitle{Temporal Context Conditioning for Seasonality-Aware Precipitation Nowcasting of High-Intensity Rainfall}

\shortauthors{G. van Nieuwkoop, S. Mehrkanoon}

\title [mode = title]{Temporal Context Conditioning for Seasonality-Aware Precipitation Nowcasting of High-Intensity Rainfall}                      



%
\author{Gijs van Nieuwkoop}



\ead{gijsvannieuwkoop@gmail.com}




\address{Department of Information and Computing Sciences, Utrecht University, Utrecht, The Netherlands}

\author{Siamak Mehrkanoon}
\cormark[1]
\ead{s.mehrkanoon@uu.nl} 

\cortext[cor1]{Corresponding author}

\begin{abstract}
Precipitation nowcasting is increasingly being approached with deep learning models that learn directly from recent radar observations. Although such models can efficiently capture short-term precipitation motion, they often lack broader contextual information about the meteorological conditions under which rainfall develops. This paper investigates whether lightweight temporal context can improve radar-based nowcasting, particularly for high-intensity rainfall. We propose the Time-Aware Small-Attention U-Net (TA-SmaAt-UNet), which extends the core SmaAt-UNet model with temporal conditioning layers that use cyclical encodings of time-of-day and time-of-year to modulate intermediate feature representations. Experiments on KNMI radar precipitation data show that temporal conditioning is most beneficial for rare, high-intensity precipitation events, while also improving the representation of seasonal variability and predicted rainfall-intensity distributions. A layer conductance analysis further indicates that the added temporal conditioning layers are actively used by the model despite their small parameter cost. These findings suggest that simple, physically motivated temporal context can improve the realism and reliability of deep learning-based precipitation nowcasts. The implementation of our models and training setup is available on \href{https://github.com/gijsvn/TA-SmaAt-UNet}{GitHub}.
\end{abstract}



\begin{keywords}
Precipitation Nowcasting \sep 
Deep Learning \sep
Temporal Conditioning \sep
Extreme Precipitation \sep
Seasonal Variability \sep
SmaAt-UNet
\end{keywords}

\maketitle

\section{Introduction}

In current weather forecasting, most state-of-the-art models stem from the paradigm of numerical weather prediction (NWP) \cite{zangl2015icon, dowell2022high, brotzge2023challenges}. Such models generate their predictions by numerically solving physical equations between different weather variables, given their initial conditions. Though reasonably accurate, such models suffer from a number of problems. Most notably, the prediction generation process of NWP models requires a lot of time and computation. This computational burden, together with latency from data assimilation and model spin-up, can limit the usefulness of NWP for very short lead-time forecasting, where predictions must be updated frequently as new observations arrive \cite{prudden2020review}.

Therefore, in recent years, much effort has gone into developing weather nowcasting models, and precipitation models in particular, which are able to rapidly generate short-horizon predictions that can be updated at an instant, taking into account the latest weather developments. Specifically, many recent advances in precipitation nowcasting have been made by adopting models from the field of deep learning \cite{shi2015convlstm, shi2017trajgru, mehrkanoon2019deep, ayzel2020rainnet, abdellaoui2021symbolic, stanczyk2021deep, trebing2021smaatunet, trebing2020wind, yang2022aa, aykas2021multistream, kaparakis2023wf}. 

These models contrast with classical NWP models by taking a data-driven approach, in which a mapping function from past observations to future observations is learned automatically from large quantities of historical data. In precipitation nowcasting, this mapping is often learned primarily from precipitation maps, allowing models to capture the short-term motion, growth, and decay of rainfall structures \cite{fernandez2021broad, trebing2021smaatunet, zhang2023nowcastnet, yin2024precipitation, vatamany2025graph, reulen2024ga}. However, by relying solely on recent precipitation observations, such models lack access to broader contextual information that influences how rainfall evolves. This can be especially relevant for high-intensity events, whose occurrence and development are affected not only by the recent spatial structure of precipitation, but also by the meteorological conditions under which they arise \cite{dai2001global_seasonal, dai2001global_diurnal}.

Motivated by known seasonal and diurnal variability in precipitation processes, this paper addresses the lack of temporal awareness in SmaAt-UNet by incorporating temporal context into the model for precipitation nowcasting \cite{trebing2021smaatunet}.Specifically, this paper proposes a temporal conditioning mechanism that enables the model to use time-of-day and time-of-year features when learning internal representations. By making this information explicitly available to the core SmaAt-Unet model, the proposed approach aims to improve nowcasting performance, particularly for high-intensity rainfall events.

This paper is organized as follows. Section~\ref{sec:related work} presents an overview of related works. Subsequently, the proposed model is presented in Section~\ref{sec:methods}. Section~\ref{sec:experiments} outlines the experimental setup used to evaluate the tested models, after which the results from these experiments are outlined in Section~\ref{sec:results}. Finally, conclusions are drawn in Section~\ref{sec:conclusion}.

\section{Related work}\label{sec:related work}

While many early deep-learning approaches formulated precipitation nowcasting mainly as a spatiotemporal sequence prediction problem from past precipitation maps, later work has increasingly recognized that recent precipitation maps provide only a partial view of the atmospheric state. ConvLSTM and TrajGRU introduced recurrent convolutional mechanisms for learning the motion and evolution of radar echoes from image sequences \citep{shi2015convlstm,shi2017trajgru}, while convolutional encoder--decoder models such as RainNet \cite{ayzel2020rainnet} and SmaAt-UNet \cite{trebing2021smaatunet} have shown that UNet-style architectures can be effective and computationally efficient for radar-based nowcasting. Transformer-based approaches have also been explored, with EarthFormer introducing a cuboid-attention-based space-time transformer for Earth-system forecasting \citep{gao2022earthformer}. More recent generative and physics-informed models, including DGMR and NowcastNet, have improved the realism and sharpness of predictions, but still primarily operate on past radar maps, optionally augmented with learned or imposed physical evolution mechanisms rather than broader contextual variables \citep{ravuri2021skilful,zhang2023nowcastnet}. This means that much of the nowcasting literature has focused on improving how models extrapolate and transform observed precipitation, while giving comparatively less direct attention to external factors that influence precipitation development but are not directly visible in the recent precipitation history.

A related line of work has attempted to address this limitation by injecting additional contextual information from other data sources. MetNet, for example, uses radar and satellite inputs together with geographic and temporal features, including longitude, latitude, elevation, hour, day, month, and forecast lead time \citep{sonderby2020metnet}. MetNet-2 extends this idea toward longer-range precipitation forecasting by combining radar, optical satellite imagery, assimilation features from HRRR, geospatial information, and explicit lead-time conditioning throughout the network \citep{espeholt2022deep}. Other studies have explored more targeted forms of data fusion: MSDM combines radar and satellite data for short-term echo prediction \citep{li2021msdm}, ASOC augments radar-based models with sparse ground-station observations and models their temporal and contextual relationships \citep{ko2022deep}, and multi-hazard thunderstorm nowcasting models use radar, lightning detections, satellite imagery, NWP variables, and elevation data to predict hazards such as lightning, hail, and heavy precipitation \citep{leinonen2023thunderstorm}. These works indicate that auxiliary context can be useful, especially for situations such as convective initiation, growth, and decay, where past precipitation maps alone may contain insufficient information about the atmospheric state.

Among possible sources of contextual information, temporal context is particularly attractive because it is universally available and does not require additional meteorological fields, satellite imagery, or NWP assimilation variables. From a meteorological perspective, similar spatial precipitation patterns may have different implications depending on when they occur. A given rainfall morphology may reflect different atmospheric conditions during the day than at night, or in summer than in winter. Diurnal and seasonal cycles are linked to factors such as surface heating, moisture availability, and the types of weather systems that are more common at different times of day or year \cite{dai2001global_diurnal, dai2001global_seasonal}. Consequently, time-of-day and time-of-year can provide useful contextual information that is not directly present in radar maps themselves, while adding almost no data-collection burden.


In spite of this, most current deep learning nowcasting models do not exploit temporal context information. Their inputs are usually limited to sequences of recent precipitation maps, which support short-term extrapolation of observed motion and intensity changes but provide no direct mechanism for using diurnal or seasonal regularities. A partial exception is MetNet, which includes calendar-like variables such as hour, day, and month as additional input feature maps \citep{sonderby2020metnet}. However, this represents an early-fusion use of temporal metadata, whereas the present work uses temporal information to condition the model’s learned feature hierarchy. Moreover, MetNet demonstrates the use of calendar variables within a broader multimodal forecasting system that also uses satellite, geospatial, and other contextual inputs. This leaves open the more targeted question addressed here: whether seasonal and diurnal context can improve a compact radar-only nowcasting model when introduced through a dedicated conditioning mechanism. To this end, the proposed method introduces a lightweight temporal conditioning mechanism into SmaAt-UNet, using cyclical encodings of time-of-day and time-of-year to modulate intermediate feature maps through channel-wise scaling and shifting, in the spirit of feature-wise conditioning layers \citep{perez2018film}. In this way, the method allows us to isolate the value of seasonal and diurnal context while preserving the radar-based nowcasting setup and the computational efficiency of the original SmaAt-UNet architecture.

\section{Methods}\label{sec:methods}

\subsection{Proposed Model: TA-SmaAt-UNet}

\begin{figure*}[t]
	\centering
    \includegraphics[width=\textwidth]{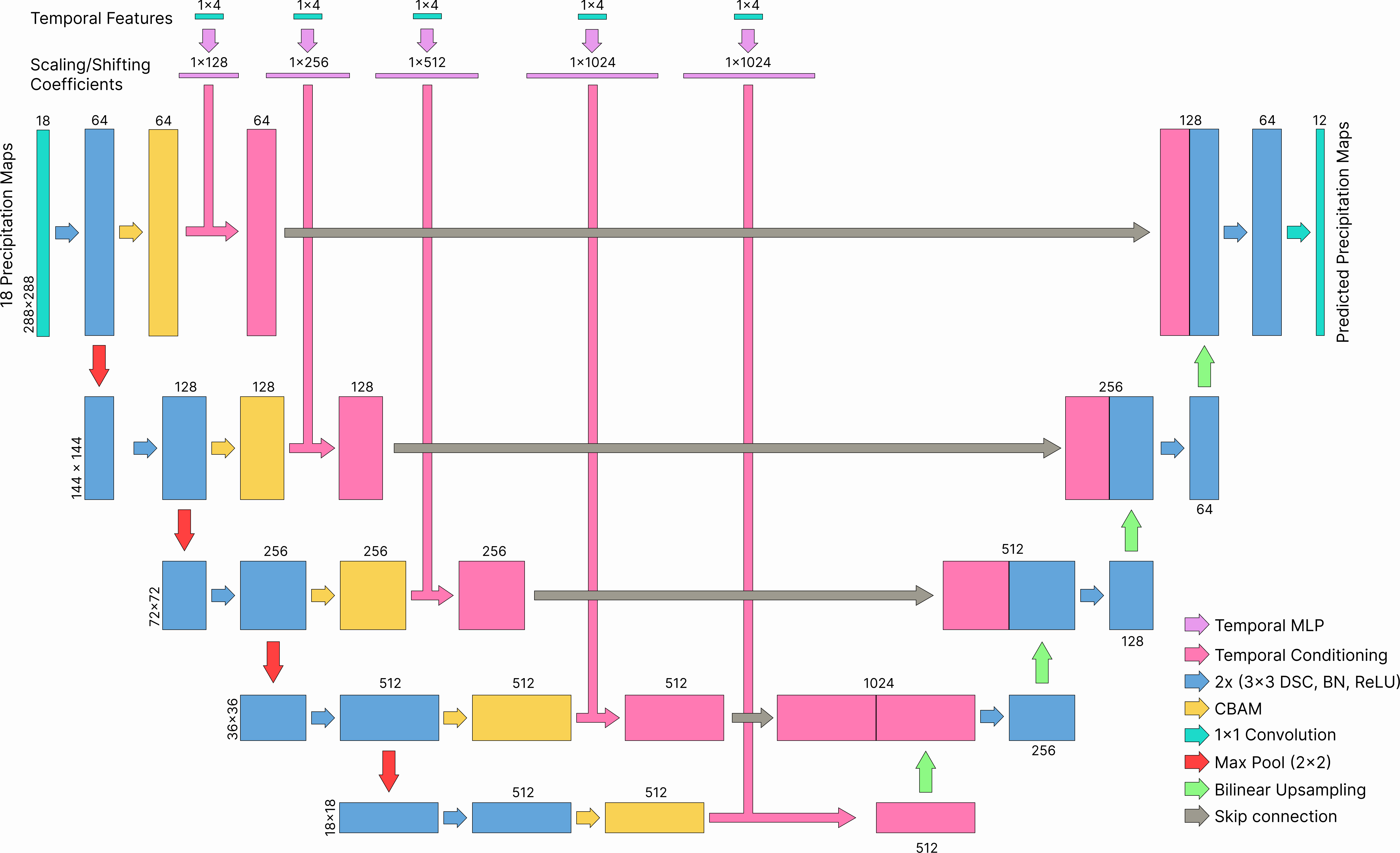}
	\caption{A schematic overview of the proposed TA-SmaAt-UNet model architecture.}
	\label{fig:architecture}
\end{figure*}

The proposed Time-Aware Small-Attention U-Net (TA-SmaAt-UNet) extends the original SmaAt-UNet core model by introducing a temporal conditioning mechanism, which uses cyclical encodings of time-of-day and time-of-year to modulate internal representations of precipitation patterns. Importantly, this modification was intended as a targeted intervention rather than as a complete redesign of the original model. As a result, the SmaAt-UNet encoder-decoder core structure is preserved, with only intermediate feature maps at the various stages of the encoder being reweighted using contextual temporal information. Figure~\ref{fig:architecture} illustrates how this mechanism is integrated in the core model architecture.

\subsubsection{Temporal Conditioning Mechanism}

The proposed modification introduces explicit temporal conditioning layers throughout the multi-level feature hierarchy. The conditioning mechanism takes a compact temporal feature vector $\mathbf{t} \in \mathbb{R}^{4}$ as input, which encodes contextual information about the time-of-day and time-of-year of the nowcasting target. More specifically, $\mathbf{t}$ is constructed using $\tau_{\mathrm{d}} \in [0,1)$ and $d_{\mathrm{y}} \in \{1,\ldots,N_{\mathrm{y}}\}$, where $N_{\mathrm{y}}$ denotes the number of days in the target year. Here, $\tau_{\mathrm{d}}$ denotes the fraction of the day elapsed at the target time, computed as follows:
\begin{equation}
\tau_{\mathrm{d}}=\frac{3600h+60m+s}{86400},
\end{equation}
where $h$, $m$, and $s$ are the hour, minute, and second of the target timestamp. Similarly, $d_{\mathrm{y}}$ denotes the day of the year, so that January 1 corresponds to $d_{\mathrm{y}}=1$ and December 31 corresponds to $d_{\mathrm{y}}=365$ or $d_{\mathrm{y}}=366$, depending on whether or not the target year is a leap year. To obtain a cyclical representation, these temporal quantities are first converted into angular variables,
\begin{equation}
\theta_{\mathrm{d}} = 2\pi \tau_{\mathrm{d}},
\qquad 
\theta_{\mathrm{y}} = 2\pi \frac{d_{\mathrm{y}}}{N_{\mathrm{y}}},
\end{equation}
after which the temporal feature vector is defined as
\begin{equation}
\mathbf{t}
=
\begin{bmatrix}
\sin (\theta_{\mathrm{d}}), \;
\cos (\theta_{\mathrm{d}}), \;
\sin (\theta_{\mathrm{y}}), \;
\cos (\theta_{\mathrm{y}})
\end{bmatrix}
\in \mathbb{R}^{4}.
\label{eq:temp_features}
\end{equation}
This encoding avoids artificial discontinuities at daily and annual boundaries and reflects the periodic nature of the underlying meteorological drivers.

At each level of the model, the temporal feature vector is transformed into channel-wise affine modulation parameters by a small level-specific densely connected network, represented in Figure~\ref{fig:architecture} as the \textit{Temporal MLP} operators. Each Temporal MLP contains a single hidden layer with 16 units, followed by an output layer whose dimensionality depends on the number of channels at that level. Specifically, for an intermediate feature map $\mathbf{X}^{(l)} \in \mathbb{R}^{C_l \times H_l \times W_l}$ at level $l$, where $C_l$ is the number of channels and $H_l \times W_l$ is the spatial resolution, the corresponding Temporal MLP is defined as follows:
\begin{equation}
f_l: \mathbb{R}^{4} \rightarrow \mathbb{R}^{2C_l},
\qquad
f_l(\mathbf{t}) = [\boldsymbol{\gamma}^{(l)}, \boldsymbol{\beta}^{(l)}],
\label{eq:temporal_mlp}
\end{equation}
where $\boldsymbol{\gamma}^{(l)}, \boldsymbol{\beta}^{(l)} \in \mathbb{R}^{C_l}$. The output layer of $f_l$ therefore contains $2C_l$ units: the first $C_l$ outputs are interpreted as channel-wise scaling coefficients $\boldsymbol{\gamma}^{(l)}$, while the remaining $C_l$ outputs are interpreted as channel-wise shifting coefficients $\boldsymbol{\beta}^{(l)}$. Thus, each channel receives one scale and one shift value, which are applied uniformly across all $H_l \times W_l$ spatial positions of that channel. More explicitly, the conditioned feature map $\mathbf{X}'^{(l)} \in \mathbb{R}^{C_l \times H_l \times W_l}$ is computed as follows:
\begin{equation}
X'^{(l)}_{c,h,w}
=
\left(1 + \gamma^{(l)}_c\right) X^{(l)}_{c,h,w}
+
\beta^{(l)}_c,
\label{eq:time_conditioning_index}
\end{equation}

for $c=1,\ldots,C_l$, $h=1,\ldots,H_l$, and $w=1,\ldots,W_l$. Here, the additive identity term in $(1+\gamma^{(l)}_c)$ is used to make $\gamma^{(l)}_c=0$ preserve the original scale of channel $c$ rather than suppressing it. This level-specific channel-wise scaling and shifting mechanism is indicated in Figure~\ref{fig:architecture} by the \textit{Temporal Conditioning} operators.

This conditioning mechanism allows the activation strength and baseline level of each learned feature channel to vary as a function of temporal context. In doing so, the model can adapt its internal representation of radar-derived precipitation patterns according to cyclical time-of-day and time-of-year information, while preserving the original core SmaAt-UNet architecture apart from the added conditioning layers.

\subsection{Other models}

To benchmark the proposed TA-SmaAt-UNet and isolate the effect of temporal conditioning, several additional models were evaluated. First, an unmodified implementation of the original SmaAt-UNet was included, providing the most direct comparison for evaluating the contribution of the proposed temporal conditioning mechanism. Second, RainNet and EarthFormer were implemented as established reference architectures for precipitation nowcasting \cite{ayzel2020rainnet, gao2022earthformer}. RainNet was included as a higher-capacity convolutional baseline, conceptually related to SmaAt-UNet and TA-SmaAt-UNet, while EarthFormer was included to compare their performance against a conceptually different attention-based forecasting architecture. Both of these models were only modified to in order to adapt their inputs and outputs to the current forecasting task. The RainNet was trained using an Adam optimizer with an initial learning rate of $10^{-4}$ and a batch size of 32, and the EarthFormer was trained using an AdamW optimizer with an initial learning rate of $2\times10^{-4}$ and a batch size of 32. Finally, a persistence model, which repeats the most recent input frame as the forecast, was included as a non-learning baseline to assess whether the deep learning models provide predictive skill beyond this simple heuristic.

\subsection{Model training}

For the sake of achieving a fair performance comparison, the training regime used for the TA-SmaAt-UNet mimicked that of the original SmaAt-UNet \cite{trebing2021smaatunet}. This meant that the model was optimized using a pixel-wise mean squared error (MSE) loss, a batch size of 16, an Adam optimizer with initial learning rate set to 0.001 and a learning rate scheduler that reduced the learning rate with a reduction factor of 0.1 whenever the validation loss did not decrease in the 4 most recent epochs. To determine the length of training, an early stopping criterion was employed, terminating training when the validation loss did not decrease over the 15 most recent epochs. For each model, five training runs were performed, after which the weights from the run with the lowest validation loss were select for evaluation on the test set. This procedure was followed in order mitigate negative effects of suboptimal weight initializations. Training was performed on a single NVIDIA H100 GPU with 94 GB of VRAM.

\subsection{Model evaluation}

To assess the model’s regression performance, we used MSE, which also served as the optimization objective during training. This metric quantifies how close the rain intensity of each pixel in the predicted precipitation maps was to the actual rain intensity of the same pixel in the associated radar precipitation image. Eq. (\ref{eq:mse}) shows the formal definition of this metric, given a predicted precipitation map $\hat{y}_i$ and a radar precipitation image $y_i$.
\begin{equation}
\mathrm{MSE} = \frac{\sum_{i=1}^{n} (y_i - \hat{y}_i)^2}{n}.
\label{eq:mse}
\end{equation}
Additionally, classification metrics can be computed by converting both predictions and radar observations of precipitation into binary maps by applying a threshold on the precipitation rate. From these binary maps, true positives (TP), true negatives (TN), false positives (FP) and false negatives (FN) can be counted. These counts can subsequently be used to compute the critical success index (CSI, Eq. (\ref{eq:csi})), probability of detection (POD, Eq. (\ref{eq:pod})), false alarm ratio (FAR, Eq. (\ref{eq:far})) and the Matthews correlation coefficient (MCC, Eq. (\ref{eq:mcc})). Computing these metrics for higher thresholds on the precipitation rate yields insights into the generalization performance of different models in increasingly rare but critical cases of heavy and extreme precipitation.
\begin{align}
\text{CSI} &= \frac{\mathrm{TP}}{\mathrm{TP}+\mathrm{FP}+\mathrm{FN}}\label{eq:csi} \\[6pt]
\text{POD} &= \frac{\mathrm{TP}}{\mathrm{TP}+\mathrm{FN}}\label{eq:pod} \\[6pt]
\text{FAR} &= \frac{\mathrm{FP}}{\mathrm{TP}+\mathrm{FP}}\label{eq:far} \\[6pt]
\text{MCC} &= 
\frac{\mathrm{TP}\,\mathrm{TN}-\mathrm{FP}\,\mathrm{FN}}
{\sqrt{(\mathrm{TP}+\mathrm{FP})(\mathrm{TP}+\mathrm{FN})(\mathrm{TN}+\mathrm{FP})(\mathrm{TN}+\mathrm{FN})}}\label{eq:mcc}.
\end{align}
CSI is often used as the main performance metric in precipitation nowcasting since it quantifies the overlap between predicted and observed rainfall events while ignoring true negatives, making it particularly suitable for imbalanced precipitation maps. POD measures the fraction of observed rainfall events that are correctly detected, reflecting the model’s sensitivity, whereas FAR quantifies the proportion of predicted rainfall events that are false alarms, characterizing overforecasting behavior. MCC provides a balanced overall measure by incorporating all four elements of the confusion matrix and remains informative under strong class imbalance. Taken together, these four metrics ought to cover the different practically important aspects of evaluating model performance.

\section{Experiments}\label{sec:experiments}

In order to make a fair performance comparison with the original SmaAt-UNet publication, the experimental setup that was used closely mimicked that of \cite{trebing2021smaatunet}. In order to train and evaluate the various models, a precipitation dataset from the Royal Netherlands Meteorological Institute (Koninklijk Nederlands Meteorologisch Instituut, KNMI) was used. The precipitation maps within this dataset were generated using two C-band Doppler weather radar stations situated in De Bilt (52.10 \textdegree N, 5.18 \textdegree E, 44 m MSL) and Den Helder (52.96 \textdegree N, 4.79 \textdegree E, 51 m MSL), the Netherlands. In total, this dataset contained roughly 420,000 precipitation maps covering region surrounding the Netherlands, spaced at 5 minute intervals over the years 2016 up until 2019. Individual precipitation maps were represented as $288 \times 288$ grids, with individual pixels covering one square kilometer, and pixel values corresponding to the total accumulated precipitation over the last 5 minutes, in millimeters. This dataset was split into a training set, covering the years from 2016 to 2018, and a testing set, which included all samples from 2019. The individual pixel values in these data sets were normalized by dividing by the maximum pixel value in the training data. From the training set, 10\% of samples was used as validation data during training. Validation samples were chosen by grouping the data into bins, using the time of the year in which samples occurred, and subsequently randomly selecting the 10\% of samples within each bin. This procedure was followed in order to ensure a stratified split over different times of the year, which was estimated to be especially important for models utilizing temporal conditioning.

The specific prediction task that was used in model development was to construct the precipitation maps from 5 minutes until 60 minutes into the future, given the precipitation maps of the most recent 90 minutes. This meant that model input was made up of 18 $288 \times 288$ precipitation maps, spaced at 5 minute intervals, and model output consisted of 12 $288 \times 288$ precipitation maps. Finally, the data was filtered to only include pairs of model input and output where the final target output precipitation map contained non-zero values in at least 50\% of pixels, as this helped reduce the total amount of samples to a more manageable level, reduce potential biases due to highly frequent zero values and increase the degree to which data reflected potential practical use cases. This left a total of 5,734 training samples and 1,557 testing samples.

\section{Results and discussion}\label{sec:results}

Table~\ref{tab:result_table} summarizes the quantitative performance of the evaluated models across the full forecasting horizon. These results indicate that RainNet obtained slightly superior MSE values compared to TA-SmaAt-UNet, whilst the MSE values of SmaAt-UNet and TA-SmaAt-UNet are comparable. However, the classification metrics for event-based precipitation prediction indicate the strenghts of TA-SmaAt-UNet more clearly. At the lowest precipitation threshold of 0.5 mm/h, the differences between the models are relatively small, but the proposed TA-SmaAt-UNet already obtains the strongest performance on the balanced CSI and MCC metrics. This indicates that the temporal conditioning mechanism improves the overall correspondence between predicted and observed rainy regions, even in the relatively common low-intensity precipitation regime. 

\begin{table*}[ht]
\centering
\small
\caption{Overview of performance metrics for the different models, averaged over forecasting time horizon.}
\label{tab:result_table}
\begin{adjustbox}{center, max width=\textwidth}
\begin{tabular}{l c c c c c c c}
\toprule
\textbf{Threshold}   & 
\textbf{Model}     & \textbf{MSE ↓}     & \textbf{CSI ↑}    & \textbf{POD ↑}    & \textbf{FAR ↓}    & \textbf{MCC ↑}    & \textbf{\# Parameters} \\
\midrule
\multirow{5}{*}{$\geq 0.5$ mm/h}
& Persistence      & 0.0301             & 0.489             & 0.617             & \underline{0.309} & 0.502             & -                      \\
& RainNet          & \textbf{0.0146}    & 0.584             & \underline{0.837} & 0.345             & 0.589             & 27,924,780             \\
& EarthFormer      & 0.0157             & 0.588             & 0.785             & \textbf{0.306}    & 0.598             & 4,113,465              \\
& SmaAt-UNet       & \underline{0.0151} & \underline{0.594} & \textbf{0.837}    & 0.330             & \underline{0.605} & 4,035,140              \\
& TA-SmaAt-UNet    & 0.0151             & \textbf{0.597}    & 0.826             & 0.322             & \textbf{0.608}    & 4,085,588              \\
\midrule
\multirow{5}{*}{$\geq 10$ mm/h}
& Persistence      & -                  & \underline{0.088} & \textbf{0.153}    & 0.844             & \underline{0.151} & -                      \\
& RainNet          & -                  & 0.085             & 0.115             & 0.764             & 0.150             & -             \\
& EarthFormer      & -                  & 0.050             & 0.057             & 0.783             & 0.108             & -             \\
& SmaAt-UNet       & -                  & 0.059             & 0.073             & \underline{0.756} & 0.126             & -              \\
& TA-SmaAt-UNet    & -                  & \textbf{0.103}    & \underline{0.140} & \textbf{0.741}    & \textbf{0.183}    & -              \\
\midrule
\multirow{5}{*}{$\geq 20$ mm/h}
& Persistence      & -                  & 0.038             & 0.071             & 0.932             & 0.069             & -                      \\
& RainNet          & -                  & \underline{0.047} & \underline{0.066} & 0.874             & \underline{0.086} & -                      \\
& EarthFormer      & -                  & 0.019             & 0.020             & \underline{0.849} & 0.053             & -                      \\
& SmaAt-UNet       & -                  & 0.039             & 0.053             & 0.867             & 0.078             & -                      \\
& TA-SmaAt-UNet    & -                  & \textbf{0.066}    & \textbf{0.094}    & \textbf{0.836}    & \textbf{0.118}    & -                      \\
\bottomrule
\end{tabular}
\end{adjustbox}
\end{table*}

The advantage of TA-SmaAt-UNet becomes more pronounced at higher precipitation thresholds. At 10 mm/h, the proposed model achieves the highest CSI, lowest FAR, and highest MCC among all evaluated models. At the 20 mm/h threshold, this pattern becomes even clearer: TA-SmaAt-UNet obtains the best score on all reported threshold-based metrics. These results suggest that temporal conditioning is particularly beneficial for rare, high-intensity rainfall events. This is consistent with the motivation of the proposed method: when precipitation becomes more extreme, the recent radar sequence alone may provide insufficient information, and the seasonal or diurnal context in which a rainfall pattern occurs can become more informative for predicting its subsequent development.

\begin{figure*}
	\centering
    \adjustbox{center}{\includegraphics[width=\textwidth]{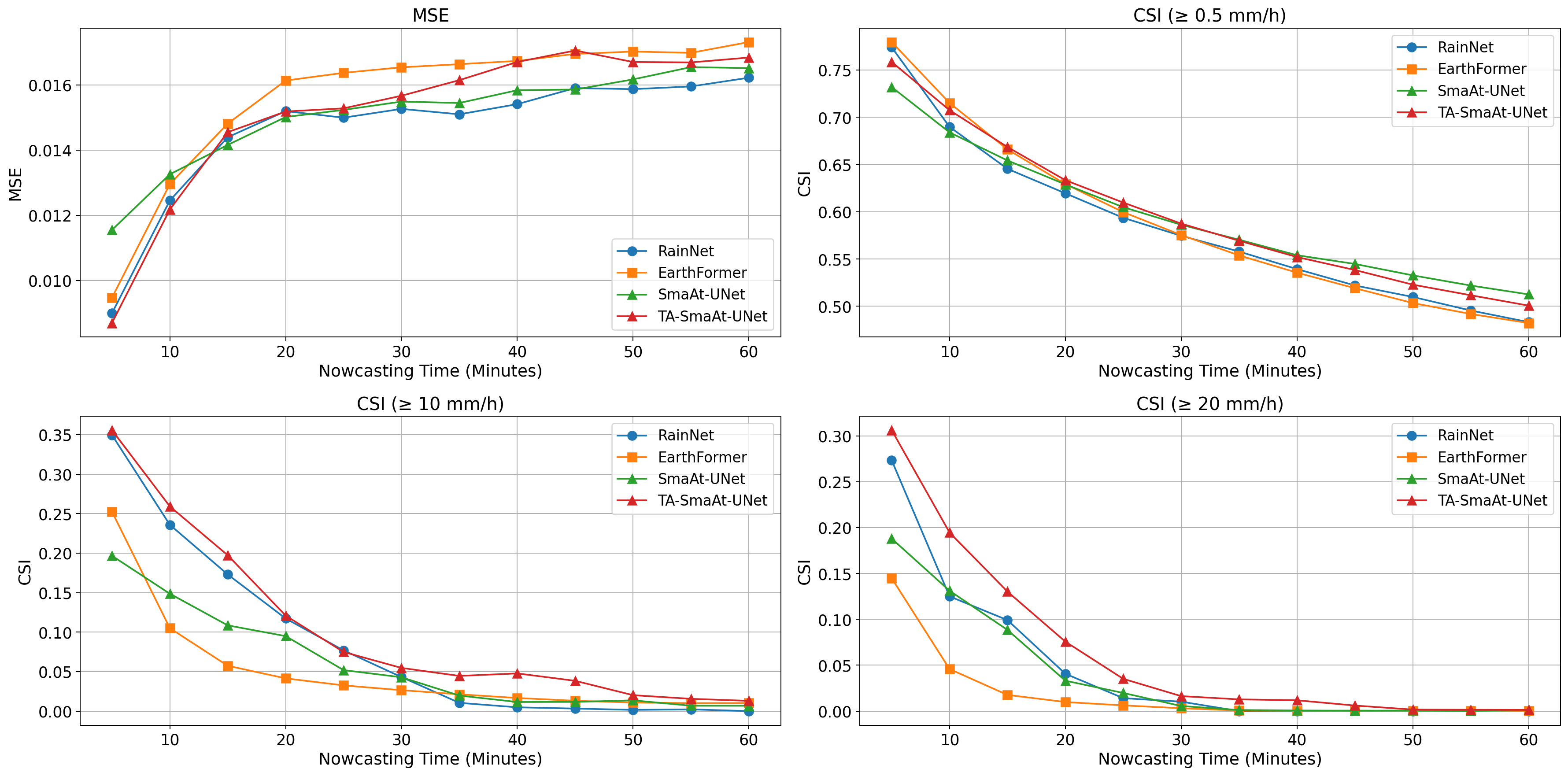}}
	\caption{MSE and CSI of different models as a function of forecast lead time.}
	\label{fig:metrics_over_forecasting_time}
\end{figure*}

Figure~\ref{fig:metrics_over_forecasting_time} further shows that the improved high-threshold performance of TA-SmaAt-UNet is not limited to a specific forecast lead time. For both the 10 mm/h and 20 mm/h CSI curves, the proposed model remains competitive or superior across most of the forecasting horizon, despite the expected decline in CSI as lead time increases. This indicates that the performance gain for intense rainfall is relatively robust over forecasting time, rather than being driven only by short-term extrapolation skill. The MSE curves also show that TA-SmaAt-UNet performs especially well at the shortest lead times, where it slightly outperforms the other models. The obtained results suggest that temporal conditioning does not come at the expense of short-term pixel-wise accuracy, while providing clearer benefits for the more difficult task of predicting high-intensity precipitation.

\begin{figure}
	\centering
    \adjustbox{center}{\includegraphics[width=0.5\textwidth]{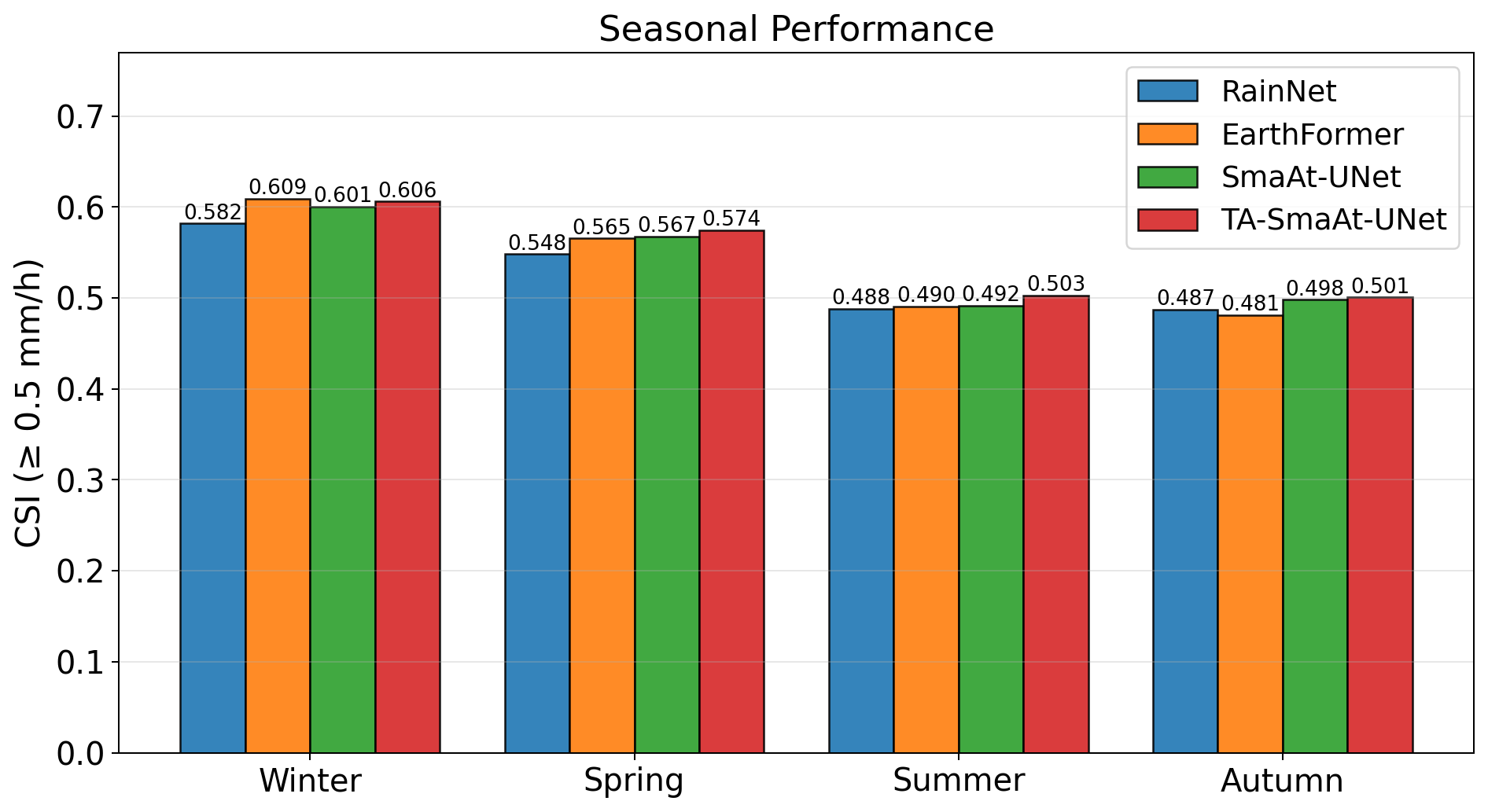}}
	\caption{Comparison of seasonal CSI scores at 0.5 mm/h precipitation threshold.}
	\label{fig:seasonal_performance}
\end{figure}

The seasonal comparison in Figure~\ref{fig:seasonal_performance} provides additional evidence that the temporal conditioning mechanism is being used in a meaningful way. TA-SmaAt-UNet improves upon the core SmaAt-UNet in every season, indicating that the benefit of temporal context is not restricted to a single part of the year. The improvement is particularly relevant in summer, which is also the most difficult season in terms of CSI. This is consistent with previous nowcasting evidence that models often struggle more with convective summer precipitation than with more persistent winter rainfall \citep{imhoff2020spatial}. From a meteorological perspective, this is plausible because summer precipitation is more often associated with showery and thunderstorm activity, which is strongly influenced by seasonal convective processes \citep{dai2001global_seasonal}. By modulating intermediate feature representations using cyclical temporal features, TA-SmaAt-UNet can therefore adapt its interpretation of similar radar patterns to the seasonal context in which they occur.

\begin{figure}
	\centering
    \adjustbox{center}{\includegraphics[width=0.5\textwidth]{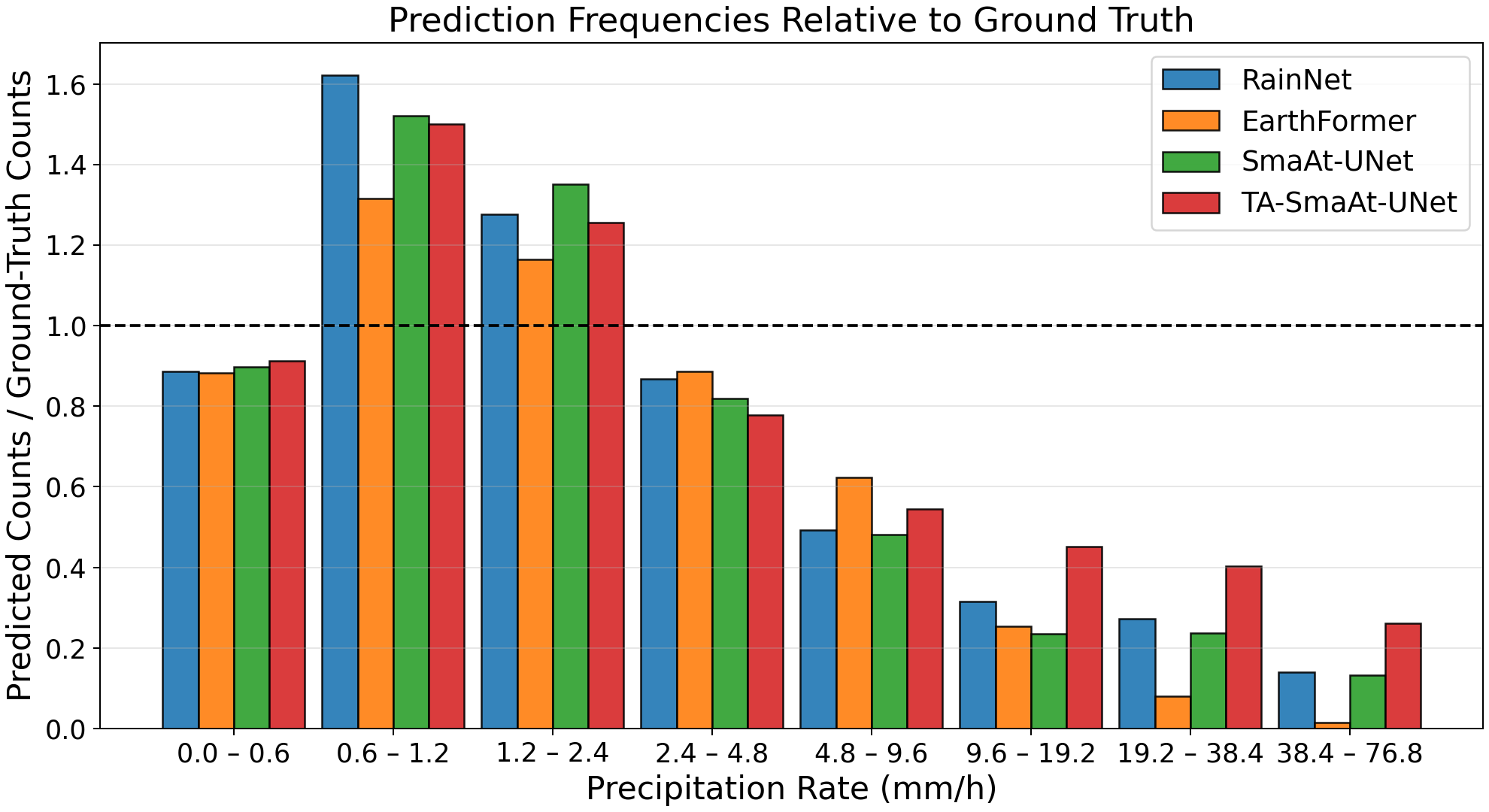}}
	\caption{Predicted precipitation-frequency distributions relative to ground truth across intensity bins.}
	\label{fig:distributions}
\end{figure}

Figure~\ref{fig:distributions} supports this interpretation from the perspective of predicted precipitation intensities. The figure shows the ratio between predicted and observed precipitation-frequency counts across intensity bins. Ideally, all bars would be close to 1, indicating that the model reproduces the ground-truth distribution of rainfall intensities. All models show some deviation from this ideal distribution, especially in the high-intensity tail, where rare rainfall events tend to be underrepresented. However, TA-SmaAt-UNet most closely follows the ground-truth distribution at both ends of the intensity range. In particular, it remains closest to the observed frequency in the highest precipitation bins, where the other models more strongly underpredict the occurrence of intense rainfall. This provides a direct explanation for the superior performance metrics at the 10 mm/h and 20 mm/h thresholds: the proposed model is better able to allocate probability mass to the high-intensity part of the distribution, rather than smoothing these events into lower rainfall intensities.

\begin{figure*}
	\centering
    \adjustbox{center}{\includegraphics[width=1.05\textwidth]{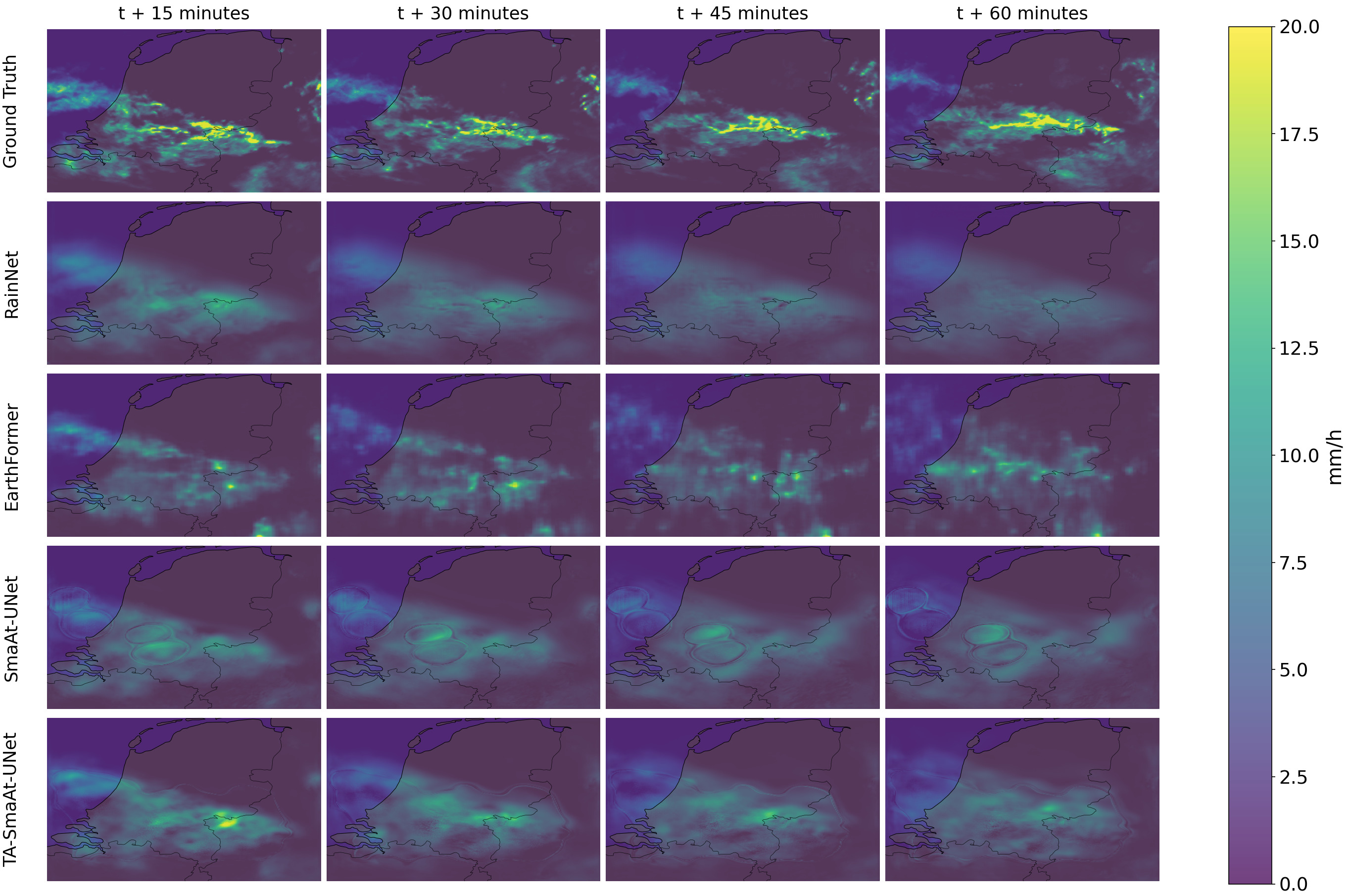}}
	\caption{Qualitative comparison of ground-truth precipitation and model predictions at selected forecast lead times.}
	\label{fig:prediction_visualization}
\end{figure*}

Finally, the qualitative comparison in Figure~\ref{fig:prediction_visualization} is consistent with the quantitative findings. RainNet produces increasingly smooth and diffuse precipitation maps at longer lead times, which reduces the sharpness of high-intensity structures. This behavior is consistent with its relatively strong MSE but weaker high-threshold performance, since spatial smoothing can reduce pixel-wise error while suppressing extremes. EarthFormer shows the opposite tendency: its predictions retain more localized structure, but become relatively patchy and spatially fragmented, with small-scale regions of rainfall that do not always correspond to coherent structures in the ground truth. SmaAt-UNet produces more coherent predictions than EarthFormer, but still shows visible artifacts and tends to underestimate some of the more intense rainfall regions. TA-SmaAt-UNet improves upon this by producing visually cleaner precipitation maps, reducing artifacts, and predicting slightly higher rainfall intensities in critical regions. These qualitative differences align with the distributional analysis and the high-threshold metrics, reinforcing the conclusion that temporal conditioning improves not only average predictive accuracy, but also the realism and intensity calibration of the predicted precipitation maps.

Overall, the results indicate that the proposed temporal conditioning mechanism provides a targeted and parameter-efficient improvement over SmaAt-UNet. The gains are modest but consistent for common low-intensity precipitation, and become substantially more important for rare high-intensity rainfall. Since these high-intensity events are often of greatest practical relevance in nowcasting applications, TA-SmaAt-UNet emerges as the most suitable model among those evaluated when the objective is not only to minimize average error, but also to preserve intense and meteorologically meaningful precipitation structures.

\subsection{Layer Conductance Analysis}

Having established the performance benefits of TA-SmaAt-UNet, we next examined whether the added temporal conditioning layers contributed meaningfully to the model's predictions. For this purpose, a layer conductance analysis was performed. Conductance attributes a model output to hidden units or layers by measuring the flow of attribution through them, extending the principle of integrated gradients from input features to internal network components \cite{dhamdhere2018important, sundararajan2017axiomatic}. For each forecast lead time, the scalar output used for attribution was the mean predicted rainfall intensity over pixels where the model predicted rain above 0.5 mm/h. Conductances were computed relative to a zero-radar baseline and several temporal baselines sampled from the training set, after which absolute conductance values were averaged. Individual layers were then grouped into four module-level categories: encoder blocks, CBAM attention modules, temporal conditioning layers (including their associated \textit{Temporal MLP}'s), and decoder blocks.

\begin{figure}
    \includegraphics[width=0.5\textwidth]{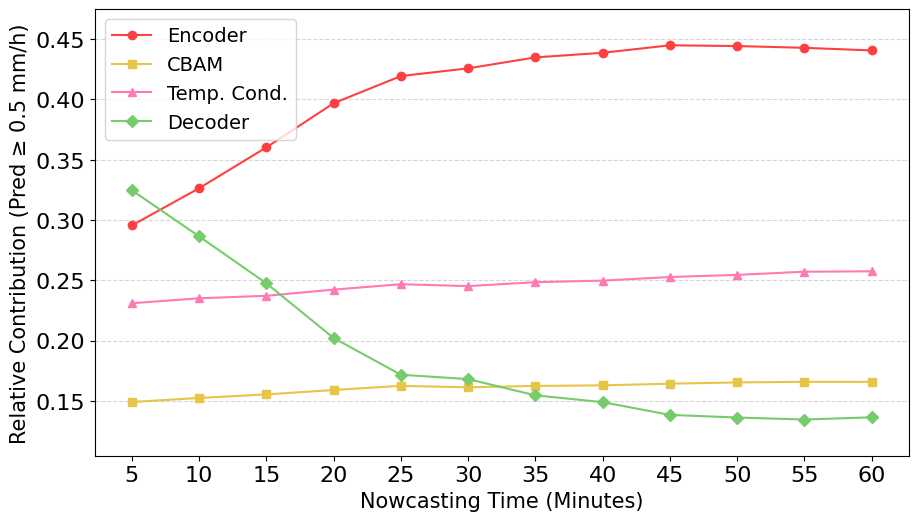}
	\caption{Relative module-level conductance of TA-SmaAt-UNet across forecast lead times for predictions above the 0.5 mm/h threshold.}
	\label{fig:conductances_over_time}
\end{figure}

Figure \ref{fig:conductances_over_time} shows a clear shift from decoder-dominated contributions at short lead times to encoder-dominated contributions at longer lead times. This pattern is consistent with short-term forecasts relying more directly on reconstructing and extrapolating visible rainfall structures, whereas longer lead times increasingly depend on more abstract encoded representations from earlier layers. More relevant to this work, the temporal conditioning contribution remains stable and increases slightly with lead time. This suggests that temporal context continues to inform the prediction process and may become more useful as forecasts depend less on immediate extrapolation alone.

Additionally, the figure also shows that the temporal conditioning layers consistently contributed more than the CBAM modules. This does not imply that temporal conditioning is generally more important than attention, since the comparison depends on the architecture, grouping strategy, attribution target, and dataset. However, within this model and attribution setup, it indicates that the temporal conditioning mechanism introduced here makes a larger relative contribution to rainy-region predictions than the pre-existing attention modules.

This interpretation is further supported by the fact that the temporal conditioning layers account for only 50,488, or approximately 1.2\% of the total trainable parameters, while their average relative conductance is approximately 24.6\%. This suggests that their contribution is not merely a consequence of module size, but that they provide a comparatively parameter-efficient mechanism for improving the model’s use of contextual information in rainfall prediction.


Together with the improved forecasting performance of TA-SmaAt-UNet, these results support the interpretation that temporal conditioning is not merely an architectural addition, but is actively used by the model when forming rainfall predictions. Although conductance analysis cannot prove that the model has learned physically correct seasonal or diurnal dynamics, it does show that temporal information is meaningfully incorporated into the prediction process. This strengthens the broader interpretation that temporal conditioning helps the model make more context-sensitive predictions, especially in settings where recent radar observations alone may provide an incomplete description of precipitation development.

\section{Conclusions}\label{sec:conclusion}




This paper introduced TA-SmaAt-UNet, a lightweight extension of SmaAt-UNet that conditions intermediate feature representations on cyclical encodings of time-of-day and time-of-year. Across the evaluated KNMI radar nowcasting task, temporal conditioning produced some improvements for common rainfall and more substantial gains for rare, high-intensity events, where the model better preserved intense precipitation structures and more closely matched the observed intensity distribution. The conductance analysis further showed that the added conditioning layers contributed consistently to the model’s predictions despite accounting for only a small fraction of the trainable parameters. These results indicate that simple temporal context can provide a parameter-efficient way to make radar-based nowcasting models more sensitive to seasonal and diurnal variability, particularly in high-impact rainfall forecasting settings.

\bibliographystyle{model1-num-names}

\bibliography{cas-refs}


\end{document}